\documentclass[12pt,a4]{article}
\usepackage[dvips]{color,graphicx}
\usepackage{bm,psfrag,afterpage,color,amsmath, amssymb,latexsym,amsthm,url}

\renewcommand{\arraystretch}{0.8}

\renewcommand{\arraystretch}{1.0}

\usepackage[subrefformat=parens]{subcaption}
\usepackage[subrefformat=parens]{caption}

\makeatletter
\def\eqnarray{\stepcounter {equation}\let \@currentlabel =\theequation
\global \@eqnswtrue
\global \@eqcnt \z@ \tabskip \@centering \let \\=\@eqncr
$$\halign to \displaywidth \bgroup \@eqnsel \hskip \@centering
$\displaystyle \tabskip \z@ {##}$&\global \@eqcnt \@ne \hfil
${\mbox{}##\mbox{}}$\hfil &\global \@eqcnt \tw@
$\displaystyle \tabskip \z@ {##}$\hfil \tabskip \@centering
&\llap {##}\tabskip \z@ \cr}
\makeatother

\renewcommand{\arraystretch}{0.9}

\begin{document}

{\baselineskip = 8mm 

\begin{center}
\textbf{\LARGE Sparse principal component regression for generalized linear models} 
\end{center}

\begin{center}
{\large Shuichi Kawano$^{1}$, \ Hironori Fujisawa$^{2,3,4}$, \\
Toyoyuki Takada$^{5}$ \ and \ Toshihiko Shiroishi$^{5}$}
\end{center}

\begin{center}
\begin{minipage}{14cm}
{
\begin{center}
{\it {\footnotesize 

\vspace{1.2mm}


$^1$ Graduate School of Informatics and Engineering,  The University of Electro-Communications, 
1-5-1 Chofugaoka, Chofu-shi, Tokyo 182-8585, Japan. \\

\vspace{1.2mm}

$^2$ The Institute of Statistical Mathematics, \\
10-3 Midori-cho, Tachikawa, Tokyo 190-8562, Japan. \\

\vspace{1.2mm}

$^3$ Department of Statistical Science, The Graduate University for Advanced Studies, \\
10-3 Midori-cho, Tachikawa, Tokyo 190-8562, Japan. \\

\vspace{1.2mm}

$^4$ Department of Mathematical Statistics, Graduate School of Medicine, Nagoya University, 
65 Tsurumai-cho, Showa-ku, Nagoya 466-8550, Japan. \\

\vspace{1.2mm}

$^5$ Mammalian Genetics Laboratory, 
National Institute of Genetics, \\ Mishima, Shizuoka 411-8540, Japan.\\

\vspace{1.2mm}

}}

\vspace{2mm}

skawano@uec.ac.jp \hspace{5mm} fujisawa@ism.ac.jp \\
ttakada@nig.ac.jp \hspace{5mm} tshirois@nig.ac.jp

\end{center}



}
\end{minipage}
\end{center}

\vspace{1mm} 

\begin{abstract}
\noindent Principal component regression (PCR) is a widely used two-stage procedure: principal component analysis (PCA), followed by regression in which the selected principal components are regarded as new explanatory variables in the model. 
Note that PCA is based only on the explanatory variables, so the principal components are not selected using the information on the response variable. 
In this paper, we propose a one-stage procedure for PCR in the framework of generalized linear models. 
The basic loss function is based on a combination of the regression loss and PCA loss. 
An estimate of the regression parameter is obtained as the minimizer of the basic loss function with a sparse penalty. 
We call the proposed method sparse principal component regression for generalized linear models (SPCR-glm). 
Taking the two loss function into consideration simultaneously, SPCR-glm enables us to obtain sparse principal component loadings that are related to a response variable. 
However, a combination of loss functions may cause a parameter identification problem, but this potential problem is avoided by virtue of the sparse penalty. 
Thus, the sparse penalty plays two roles in this method. 
The parameter estimation procedure is proposed using various update algorithms with the coordinate descent algorithm. 
We apply SPCR-glm to two real datasets, doctor visits data and mouse consomic strain data. 
SPCR-glm provides more easily interpretable principal component (PC) scores and clearer classification on PC plots than the usual PCA. 
\end{abstract}

\begin{center}
\begin{minipage}{14cm}
{
\vspace{3mm}



}
\end{minipage}
\end{center}

\baselineskip = 8mm


\section{Introduction}

Principal component regression (PCR) (Massy, 1965; Jolliffe, 1982) is a widely used two-stage procedure: one first performs principal component analysis (PCA) (Pearson, 1901; Jolliffe, 2002) and next considers a regression model in which the selected principal components are regarded as new explanatory variables.  
PCR has many extensions (Hartnett {\it et al.}, 1998; Rosital \textit{et al.}, 2001; Reiss and Ogden, 2007; Wang and Abbott, 2008). 
However, we should remark that PCA is based only on the explanatory variables, so the principal components are not selected using the information on the response variable. 
If the response variable has a close relationship with the principal components having small eigenvalues, PCR cannot achieve sufficitent prediction accuracy.

To overcome this problem, Kawano \textit{et al.} (2015) proposed a one-stage procedure for PCR. 
The basic loss function for this one-stage procedure is based on a combination of the regression squared loss and PCA loss (Zou \textit{et al.}, 2006). 
The estimate of the regression parameter is obtained as the minimizer of the basic loss function with a sparse penalty. 
This proposed method is called the sparse principal component regression (SPCR). 
SPCR enables us to obtain sparse principal component loadings that are related to a response variable, because the two loss functions are simultaneously taken into consideration. 
A combination of loss functions may cause a parameter identification problem, but this problem is overcome by virtue of the sparse penalty. 
Thus, the sparse penalty plays two roles in this method. 
The parameter estimation procedure was proposed using various iterative algorithms with the coordinate descent algorithm. 
However, the response variable is restricted to a continuous variable.

In this paper, we propose a one-stage procedure for PCR in the framework of generalized linear models (McCullagh and Nelder, 1989). 
The regression loss is replaced by the negative log-likelihood function. 
The proposed method is called the sparse principal component regression for generalized linear models (SPCR-glm). 
The main difference in SPCR-glm from SPCR is the parameter estimation procedure, because the negative log-likelihood function in generalized linear models is more complex than the regression squared loss. 
To obtain the parameter estimate, we propose a novel update algorithm combining various ideas with the coordinate descent algorithm (Friedman {\it et al.}, 2007; Wu and Lange, 2008).


We apply SPCR-glm to two real datasets, doctor visits data and mouse consomic strain data, with a Poisson regression model and multi-class logistic model, respectively. 
SPCR-glm provides more easily interpretable principal component (PC) scores and clearer classification on PC plots than the usual PCA. 
For the doctor visits data, we can also obtain more clearly interpretable PC scores. 
For the consomic strain mouse data, we can also extract characteristic mouse consomic strains with smaller within-variance.

This paper is organized as follows. 
In Section \ref{preliminaries}, we review sparse principal component analysis (SPCA) and SPCR. 
In Section \ref{SPCRforGLM}, we propose SPCR-glm and introduce some special cases. 
In Section \ref{implementation}, we provide a parameter estimation procedure for SPCR-glm and discuss the selection of tuning parameters. 
Monte Carlo simulations and real data analyses are illustrated in Sections \ref{IllustrativeExample}, \ref{NumericalStudy}, and \ref{Applications}. 
Concluding remarks are given in Section \ref{Concluding}. 
The R language software package {\tt spcr}, which implements SPCR-glm, is available on the Comprehensive R Archive Network (R Core Team, 2016). 
Supplementary materials can be found at \url{https://sites.google.com/site/shuichikawanoen/research/suppl_spcr-glm.pdf}.

\section{Preliminaries}
\label{preliminaries}

\subsection{Sparse principal component analysis}
\label{SPCA}
Let $X=({\bm x}_1, \ldots, {\bm x}_n)^T$ be an $n \times p$ data matrix with $n$ observations and $p$ variables. 
Without loss of generality, the columns of the matrix $X$ are assumed to be centered. 
PCA is formulated as the following least squares problem (e.g., Hastie {\it et al.}, 2009): 
\begin{eqnarray}
\min_{B} \sum_{i=1}^n || {\bm x}_i - B B^T {\bm x}_i ||^2_2 \ \ \ \  {\rm subject \ to} \ \ \ B^T B = I_{k}, 
\label{PCALeastLoss}
\end{eqnarray}
where $B=({\bm \beta}_1,\ldots,{\bm \beta}_k)$ is a $p \times k$ principal component loading matrix, $k$ denotes the number of principal components, $I_k$ is the $k \times k$ identity matrix, and $|| \cdot ||_2$ is the $L_2$ norm defined by $|| {\bm z} ||_2 = \sqrt{{\bm z}^T {\bm z}}$ for an arbitrary finite vector $\bm z$. 
Let $X=U D V^T$, where $U$ is an $n \times p$ matrix with $U^T U = I_p$, $V=({\bm v}_1,\ldots,{\bm v}_p)$ is a $p \times p$ orthogonal matrix, and $D = {\rm diag} (d_1,\ldots,d_p)$ is a $p \times p$ matrix with $d_1 \geq \cdots \geq d_p \geq 0$. 
Then, the estimate of $B$ is given by $V_k Q^T$, where $V_k = ({\bm v}_1,\ldots,{\bm v}_k)$ and $Q$ is an arbitrary $k \times k$ orthogonal matrix.

To easily interpret the principal component loading matrix $B$, Zou \textit{et al.} (2006) proposed SPCA, which is given by
\begin{eqnarray}
\min_{A, B}  \left\{ \sum_{i=1}^n || {\bm x}_i - A B^T {\bm x}_i ||^2_2 + \lambda \sum_{j=1}^k || {\bm \beta}_j ||^2_2 + \sum_{j=1}^k \lambda_{1,j} || {\bm \beta}_j ||_1 \right\} \ \ \ {\rm subject \ to} \ \  A^T A = I_{k}, 
\label{spca}
\end{eqnarray}
where $A=({\bm \alpha}_1,\ldots,{\bm \alpha}_k)$ is a $p \times k$ matrix, $\lambda$ and the $\lambda_{1,j}$'s \ $(j=1,\ldots,k)$ are non-negative regularization parameters, and $|| \cdot ||_1$ is the $L_1$ norm defined by $|| \bm z ||_1 = \sum_{j=1}^p |z_j|$ for an arbitrary finite vector $\bm z =(z_1,\ldots,z_p)^T$. 
A simple calculation shows that SPCA can be expressed as
\begin{eqnarray*}
\min_{A, B} \sum_{j=1}^k  \left\{ || X {\bm \alpha}_j - X {\bm \beta}_j  ||^2_2 + \lambda || {\bm \beta}_j ||^2_2 +  \lambda_{1,j} || {\bm \beta}_j ||_1 \right\} \ \ \ {\rm subject \ to} \ \ \ A^T A = I_{k}. 
\end{eqnarray*}
Given a fixed $B$, the minimizer $A$ is obtained by solving the reduced rank Procrustes rotation, which is introduced in Zou \textit{et al.} (2006). 
Given a fixed $A$, the minimization problem for $B$ is consistent with that in the elastic net (Zou and Hastie, 2005), so it can be solved using the LARS algorithm (Efron {\it et al.}, 2004) or the coordinate descent algorithm (Friedman \textit{et al.}, 2007; Wu and Lange, 2008). 
The parameter estimation procedure can be proposed via an alternate update algorithm of $A$ and $B$.


\subsection{Sparse principal component regression for continuous response}

Suppose that we have continuous response variables $y_1,\ldots,y_n$ and explanatory variables ${\bm x}_1,\ldots, {\bm x}_n$. 
SPCR is a regression model with a few principal components that include the information about the response variable. 
Then SPCR is formulated as
\begin{eqnarray}
&& \min_{A, B, \gamma_0, {\bm \gamma}} \Big\{ \sum_{i=1}^n \left(y_i - \gamma_0 - {\bm \gamma}^T B^T {\bm x}_i \right)^2 + w \sum_{i=1}^n || {\bm x}_i - A B^T {\bm x}_i ||^2_2   \nonumber \\
&&  \hspace{30mm} + \lambda_{\beta} \xi \sum_{j=1}^k || {\bm \beta}_j ||^2_2 + \lambda_{\beta} (1-\xi) \sum_{j=1}^k  || {\bm \beta}_{j} ||_1 + \lambda_{\gamma} || {\bm \gamma} ||_1\Big\}  \label{spcr} \\
&& {\rm subject \ to} \ \ \ A^T A = I_{k}, \nonumber
\end{eqnarray}
where $\gamma_0$ is an intercept, ${\bm \gamma} = (\gamma_1,\ldots,\gamma_k)^T$ is a coefficient vector, $ \lambda_{\beta}$ and $\lambda_{\gamma}$ are non-negative regularization parameters, $w$ is a positive tuning parameter, and $\xi$ is a tuning parameter in $[0,1)$. 
In Formula (\ref{spcr}), the first term is the squared loss function of a linear regression model that employs the principal components $B^T {\bm x}$ as explanatory variables, and the second term  is the loss function of PCA, which is used in SPCA. 
Sparse regularizations in SPCR have two roles: estimating some parameters as zero and overcoming the parameter identification problem (for details, see Kawano \textit{et al.}, 2015). 

The minimization problem (\ref{spcr}) is a quadratic programming problem with respect to each parameter $\{ B, \gamma_0, {\bm \gamma} \}$, so that it is easy to construct the parameter estimation procedure, which was proposed using the coordinate descent algorithm by Kawano \textit{et al.} (2015).

In SPCR, the response variable is restricted to being continuous variable. 
In Section \ref{SPCRforGLM}, SPCR is extended to the framework of generalized linear models (McCullagh and Nelder, 1989) to deal with various types of data, including binary, count, and multiclass data.

\section{Sparse principal component regression for generalized linear models}
\label{SPCRforGLM}
We assume that the response variable given the explanatory variables is generated from the exponential family
\begin{eqnarray}
f(y_i | {\bm x}_i ; \theta ({\bm x}_i), \phi) = \exp \left\{ \frac{ y_i \theta ({\bm x}_i) - u(\theta ({\bm x}_i)) }{\phi} + v ( y_i, \phi ) \right\},
\label{ExponentialFamily}
\end{eqnarray}
where $\theta ({\bm x}_i)$ is a canonical parameter, $\phi$ is a nuisance parameter, and $u(\cdot)$ and $v(\cdot,\cdot)$ are known specific functions. 
The mean ${\rm E} ( Y_i ) \ (=\mu_i)$ and variance ${\rm Var} ( Y_i )$ in the distribution (\ref{ExponentialFamily}) are given by $u^\prime (\theta ({\bm x}_i))$ and $\phi u^{\prime \prime}  (\theta ({\bm x}_i))$, respectively. 
Let $\kappa_i$ be the linear predictor in the framework of generalized linear models with $\kappa_i = h( u^\prime (\theta ({\bm x}_i)) )$, where $h(\cdot)$ is a link function (McCullagh and Nelder, 1989). 
From this relationship, (\ref{ExponentialFamily}) is reformulated as
\begin{eqnarray}
f(y_i | {\bm x}_i ; \theta ({\bm x}_i), \phi) = \exp \left\{ \frac{ y_i r(\kappa_i) - s(\kappa_i) }{\phi} + v ( y_i, \phi ) \right\},
\label{ExponentialFamily2}
\end{eqnarray}
where $r (\cdot) = u^{\prime -1} \circ h^{-1} (\cdot)$ and $s (\cdot) = u \circ u^{\prime -1} \circ h^{-1} (\cdot)$. 
The link function $h (\cdot)$ is often canonical with $h (\cdot) = u^{\prime -1} (\cdot)$. 
Then we have $r(\kappa_i) = \kappa_i$.

Suppose that 
$$\kappa_i(\bm{x}_i;\gamma_0,\bm{\gamma},B)=\gamma_0 + {\bm \gamma}^T B^T {\bm x}_i,$$
where  
$\gamma_0$ is an intercept, ${\bm \gamma} = (\gamma_1,\ldots,\gamma_k)^T$ is a coefficient vector, and $B = ({\bm \beta}_1, \ldots, {\bm \beta}_k)$ is a $p \times k$ loading matrix. 
The PC score $B^T {\bm x}_i$ is regarded as a new explanatory variable. 
Here we consider the minimization problem 
\begin{eqnarray}
&& \min_{A, B,\gamma_0, {\bm \gamma}}  \Bigg[ 
L_{\rm reg}(\gamma_0,\bm{\gamma},B) + w L_{\rm PCA}(A,B) + P_1(B;\lambda_\beta,\xi) + P_2(\bm{\gamma};\lambda_\gamma) \Bigg] \label{SPCRglm} \\
&& \hspace*{20mm} {\rm subject \ to} \ \ \ A^T A = I_{k}, \nonumber
\end{eqnarray}
where
\begin{eqnarray*}
L_{\rm reg}(\gamma_0,\bm{\gamma},B) &=& - \sum_{i=1}^n \log f(y_i | {\bm x}_i ; \kappa_i(\bm{x}_i;\gamma_0,\bm{\gamma},B), \phi), \\
L_{\rm PCA}(A,B) &=&  \sum_{i=1}^n || {\bm x}_i - A B^T {\bm x}_i ||_2^2, \\
P_1(B;\lambda_\beta,\xi) &=& \lambda_\beta \xi \sum_{j=1}^k || {\bm \beta}_j ||_2^2 + \lambda_{\beta} (1-\xi) \sum_{j=1}^k   || {\bm \beta}_{j} ||_1, \\
P_2(\bm{\gamma};\lambda_\gamma) &=& \lambda_\gamma || \bm \gamma ||_1,
\end{eqnarray*}
$w$ is a {positive tuning parameter}, $A$ is a $p \times k$ matrix, $\lambda_{\beta}$ and $\lambda_{\gamma}$ are {non-negative regularization parameters}, and $\xi$ is a tuning parameter between zero and one. 
The loss function $L_{\rm reg}$ is the negative log-likelihood, $L_{\rm PCA}$ is another loss function of PCA (Zou \textit{et al.}, 2006), $P_1(B;\lambda_\beta,\xi)$ is the elastic net regularization on the loading matrix $B$, and $P_2(\bm{\gamma};\lambda_\gamma)$ is the sparse regularization penalty for $\bm{\gamma}$, which implies an automatic selection of principal components. We do not adopt the elastic net regularization for $\bm{\gamma}$, because the new explanatory variables based on $B^T \bm{x}$ are expected to be weakly correlated by virtue of the PCA loss. 
}
The tuning parameter $w$ plays a role in the weight of the PCA loss; we obtain a better prediction accuracy as $w$ becomes smaller, while we obtain a better  formulation of principal component loadings as $w$ becomes larger. 
The tuning parameter $\xi$ controls the trade-off between the $L_1$  and $L_2$ penalties for $B$ (Zou and Hastie, 2005). 
The minimization problem (\ref{SPCRglm}) enables us to perform {regression analysis and PCA} simultaneously. 
We call this procedure sparse principal component regression for generalized linear models (SPCR-glm).

In the minimization problem (\ref{SPCRglm}), there exists an identification problem for the parameters $B$ and ${\bm \gamma}$: for an arbitrary orthogonal matrix $P$, we see $\bm \gamma^T B^T = \bm \gamma^{*T} B^{*T}$, where ${\bm \gamma}^* = P {\bm \gamma}$ and $B^* = B P^T$. 
As discussed in Tibshirani (1996), Jennrich (2006), Choi \textit{et al}. (2011), and Hirose and Yamamoto (2015), this problem is overcome by sparse regularization for $B$ or $\bm \gamma$. 
The sparse regularizations for $B$ and $\bm \gamma$ in Formula (\ref{SPCRglm}), therefore, have two roles: sparse estimation and identification of parameters.

In our numerical experiments, we {encountered the problem} that SPCR-glm {sometimes failed to give many} sparse estimates of the loading matrix $B$. 
To obtain many sparse estimates of $B$, we propose assigning different regularization parameters {for the components of} $B$: the term $\lambda_{\beta} (1-\xi) \sum_{j=1}^k   || {\bm \beta}_{j} ||_1$ is replaced by $ (1-\xi) \sum_{j=1}^k \sum_{l=1}^p \lambda_{\beta, lj} | {\beta}_{lj} |$, where $\lambda_{\beta, lj}$ is a {non-negative regularization parameter}. 
We call this procedure adaptive sparse principal component regression for generalized linear models (aSPCR-glm). 
In our numerical experiments, we {utilized} $\lambda_{\beta,lj} = \lambda_\beta /| \hat{\beta}_{lj}^\dagger |^q$ {with $q \geq 0$, where} $\hat{\beta}_{lj}^\dagger$ {is} an estimate of $\beta_{lj}$ derived from SPCR-glm. 
{This idea is based on the adaptive lasso by  Zou (2006).}

\subsection{Sparse principal component logistic regression}
\label{SPCR_logistic}
Suppose that we have a binary response variable $y_i \in \{0,1\}$. 
The logistic regression model is given when $\phi=1, u (\kappa_i) = \log \{ 1 + \exp(\kappa_i) \}$, and $v(y_i,\phi)=0$. 
The regression loss function is
\begin{eqnarray*}
L_{\rm reg} = - \sum_{i=1}^n \left[ y_i ( \gamma_0 + {\bm \gamma}^T B^T {\bm x}_i ) - \log \{ 1 + \exp( \gamma_0 + {\bm \gamma}^T B^T {\bm x}_i ) \} \right].
\end{eqnarray*}

\subsection{Sparse principal component Poisson regression}
Suppose that we have a count response variable $y_i \in \{0, 1, 2, \ldots\}$. The Poisson regression model is given when $\phi=1$, $u (\kappa_i) = \exp(\kappa_i)$, and $v(y_i,\phi)= - \log y_i ! $. The regression loss function is 
\begin{eqnarray*}
L_{\rm reg} &=& - \sum_{i=1}^n \{ y_i ( \gamma_0 + {\bm \gamma}^T B^T {\bm x}_i ) - \exp( \gamma_0 + {\bm \gamma}^T B^T {\bm x}_i )   - \log y_i ! \}.
\end{eqnarray*}

\subsection{Sparse principal component multiclass-logistic regression}
\label{sec:multiclass}
We assume that $G$ levels of categorical values are observed for response variable $C$. 
The multiclass-logistic regression model is given by
\begin{eqnarray*}
{\rm Pr} (C=g | {\bm x}) = \frac{ \exp ( \gamma_{0g} + {\bm \gamma}_g^T B^T {\bm x} ) }{\sum_{g=1}^G \exp ( \gamma_{0g} + {\bm \gamma}_g^T B^T {\bm x} )}, \quad (g=1,\ldots,G),
\end{eqnarray*}
where $\gamma_{0g}$'s $(g=1,\ldots,G)$ are intercepts, and ${\bm \gamma}_g = (\gamma_{1g},\ldots, \gamma_{pg})^T$'s $(g=1,\ldots,G)$ are the coefficient vectors. 
This is not the traditional model. 
This symmetric modeling procedure was used by Zhu and Hastie (2004), because it enables us to make an easier {parameter estimation algorithm} than the traditional model. 
If we denote by $Y = ({\bm y}_1,\ldots,{\bm y}_G)$ the $n \times G$ indicator response matrix with elements $y_{ig} = I( c_i = g )$, then the regression loss function is given by
\begin{eqnarray*}
L_{\rm reg} &=& - \sum_{g=1}^G \sum_{i=1}^n [ y_{ig} (\gamma_{0g} + {\bm \gamma}_g^T B^T {\bm x}_i) - \log \{ 1 + \exp(\gamma_{0g} + {\bm \gamma}_g^T B^T {\bm x}_i) \}.
\end{eqnarray*}
We note that, by slightly modifying the density (\ref{ExponentialFamily}) with the response vectors, the minimization problem for the multiclass-logistic regression becomes a special case of SPCR-glm.

There is an identification problem in the symmetric modeling. 
The probabilities with $\{ \gamma_{0g}, {\bm \gamma}_{g} \}$ are identical to these with $\{ \gamma_{0g}-\gamma_{0}^*, {\bm \gamma}_{g} - {\bm \gamma}^* \}$. 
This cannot be overcome without imposing constrains.  
This crucial problem was discussed in Friedman \textit{et al.} (2010), according to which $\gamma_j^* \ (j=1,\ldots,p)$ is provided by a median of $\{ \gamma_{j1}, \ldots, \gamma_{jG} \}$, and then $\gamma_{0}^*$ is determined by mean centering of $\{ \gamma_{01}, \ldots, \gamma_{0G} \}$ by employing regularization. 
For details, see Theorem 1 in Friedman \textit{et al.} (2010).



\subsection{Related work}
\label{RelatedWork}
As a related work, we refer to PLS generalized linear regression (PLS-GLR) proposed by Bastien \textit{et al}. (2005). 
PLS-GLR can perform partial least squares (Wold, 1975; Frank and Friedman, 1993) when the response variable belongs to the exponential family or is censored. 
Although PLS-GLR is similar to SPCR-glm, PLS-GLR does not integrate the two loss functions for generalized linear models and PCA with the $L_1$-type regularization. 
In addition, PLS-GLR is a two-stage procedure, but SPCR-glm is a one-stage procedure. 
In Sections \ref{IllustrativeExample}, \ref{NumericalStudy}, and \ref{Applications}, we will compare the two methods numerically.

\section{Implementation}
\label{implementation}
\subsection{Computational algorithm}
Since SPCR-glm is a special case of aSPCR-glm, we focus on an estimation algorithm for aSPCR-glm. 
We estimate the parameters $\{B, \gamma_0, \bm \gamma\}$ in aSPCR-glm by the coordinate descent algorithm (Friedman {\it et al.}, 2007; Wu and Lange, 2008), because the minimization problem includes $L_1$-type regularizations. 
The parameter $A$ is estimated according to Zou \textit{et al.} (2007).

In SPCR, it is easy to implement the coordinate descent algorithm, because the optimization is a quadratic programming problem. 
However, in aSPCR-glm, the optimization is not a quadratic programming problem, because the log-likelihood function (\ref{ExponentialFamily}) is a nonlinear convex function in general. 
Therefore, for the current estimates of the parameters $\{ \tilde{B}, \tilde{\gamma}_0, \tilde{\bm \gamma} \}$, we apply second-order Taylor expansion to the negative log-likelihood function around current estimates. 
The Taylor expansion leads to the approximated minimization problem given by
\begin{eqnarray*}
&&\min_{A, B,\gamma_0, {\bm \gamma}}  \Bigg[ - \sum_{i=1}^n \eta_i (z_i - \gamma_0 - {\bm \gamma}^T B^T {\bm x}_i)^2 +  w \sum_{i=1}^n || {\bm x}_i - A B^T {\bm x}_i ||_2^2  \nonumber\\
&& \hspace{25mm}    + \ \lambda_\beta \xi \sum_{j=1}^k || {\bm \beta}_j ||_2^2 + (1-\xi) \sum_{j=1}^k \sum_{l=1}^p \lambda_{\beta, lj} | {\beta}_{lj} | + \lambda_\gamma || \bm \gamma ||_1 \Bigg] \label{SPCRglmAppro} \\
&& {\rm subject \ to} \ \ \ A^T A = I_{k}, \nonumber
\end{eqnarray*}
where 
\begin{eqnarray*}
\eta_i &=& - \frac{u^{\prime \prime} ( \tilde{\gamma}_0 + \tilde{\bm \gamma}^T \tilde{B}^T {\bm x}_i )}{2 \phi}, \\
z _i &=& \tilde{\gamma}_0 + \tilde{\bm \gamma}^T \tilde{B}^T {\bm x}_i - \frac{ y_i - u^{\prime} ( \tilde{\gamma}_0 + \tilde{\bm \gamma}^T \tilde{B}^T {\bm x}_i ) }{u^{\prime \prime} ( \tilde{\gamma}_0 + \tilde{\bm \gamma}^T \tilde{B}^T {\bm x}_i )}.
\end{eqnarray*}
This approximation leads to the updated equations given as follows. 


\begin{description}
\item[ $\beta_{lj}$ given $\gamma_0$, $\gamma_j$, and $A$:] 
The coordinate-wise update for $\beta_{lj}$ has the following form:
\begin{eqnarray}
&& \hat{\beta}_{l^{\prime} j^{\prime}} \leftarrow \frac{S \left( \sum_{i=1}^n x_{i l^{\prime}} \left\{ \eta_i Z_i \gamma_{j^{\prime}} + 2 w Y_{j^{\prime} i }^* \right\}, (1-\xi) \lambda_{\beta, l^\prime j^\prime} \right)}{ \gamma_{j^\prime}^2 \sum_{i=1}^n \eta_i x_{i l^\prime}^2 + 2 w \sum_{i=1}^n x_{i l^\prime}^2 + 2 \lambda_\beta \xi }, \label{UpdateBeta} \\
&& \hspace{80mm} (l^{\prime} = 1,\ldots,p; \ j^{\prime} = 1,\ldots,k), \nonumber
\end{eqnarray}
where
\begin{eqnarray*}
Z_i &=&  z_i - \gamma_0 - \sum_{j=1}^k \sum_{l \neq l^{\prime}} \gamma_j \beta_{lj} x_{il} - \sum_{j \neq j^{\prime}} \gamma_j \beta_{l^{\prime} j} x_{i l^{\prime}}, \\
Y^*_{j^{\prime} i} &=& y_{j^{\prime} i}^* - \sum_{l \neq l^{\prime}} \beta_{l j^{\prime}} x_{il},
\end{eqnarray*}
and $S(z, \eta)$ is the soft-thresholding operator with
\begin{eqnarray*}
{\rm sign} (z) (|z| - \eta)_+   = \left\{ \begin{array}{ll}
z - \eta, & (z > 0 \ {\rm and} \ \eta < |z|), \\
z + \eta, & (z < 0 \ {\rm and} \ \eta < |z|),\\
0, & (\eta \geq |z|). \\
\end{array} \right.
\end{eqnarray*}
Here ${\bm y}_j^* = X {\bm \alpha}_j$.

\item[ $\gamma_j$ given $\gamma_0$, $\beta_{lj}$, and $A$:] 
The update expression for $\gamma_j$ is given by
\begin{eqnarray}
\hat{\gamma}_{j^{\prime}} \leftarrow \frac{S \left( \sum_{i=1}^n \eta_i z^{**}_i x^*_{ij^{\prime}}, \lambda_{\gamma} \right) }{\sum_{i=1}^n \eta_i x^{*2}_{i j^{\prime}}}, \quad (j^{\prime} = 1,\ldots, k),
\label{UpdateGamma}
\end{eqnarray}
where
\begin{eqnarray*}
x^*_{ij} &=& {\bm \beta}^T_j {\bm x}_i, \\
y^{**}_i &=& z_i - \gamma_0 - \sum_{j \neq j^{\prime}} \gamma_j x^*_{ij}.
\end{eqnarray*}

\item[ $A$ given $\gamma_0$, $\beta_{lj}$, and $\gamma_j$:]
The estimate of $A$ is obtained by 
\begin{eqnarray*}
\hat{A} = U V^T,
\end{eqnarray*}
where $(X^T X) B = U D V^T$.

\item[ $\gamma_0$ given $\beta_{lj}$, $\gamma_j$, and $A$:]
The estimate of $\gamma_0$ is derived from 
\begin{eqnarray*}
\hat{\gamma}_0 = \frac{1}{\sum_{i=1}^n \eta_i } \sum_{i=1}^n \eta_i \left\{ z_i - \sum_{j=1}^k \hat{\gamma}_j \left( \sum_{l=1}^p \hat{\beta}_{lj} x_{il} \right) \right\}.
\end{eqnarray*}
\end{description}
The update procedure can be directly implemented for the logistic model and Poisson regression model. 
The multiclass-logistic regression model has a special structure, as described in Section~\ref{sec:multiclass}, so we need a slight modification, which is given in the supplementary material Appendix A.

The updates described earlier lead us to the parameter estimation procedure, which is summarized in the following steps:  
\begin{description}
\item[Step 1] Set the values of the tuning parameters $\{w, \xi\}$ and the regularization parameters $\{ \lambda_\beta, \lambda_{\beta,lj}, \lambda_\gamma \}$. 
\item[Step 2] Initialize the parameters $\{A, B, \gamma_0, \bm \gamma\}$. 
\item[Step 3] Update the objective $\eta_i \ (i=1,\ldots,n)$. 
\item[Step 4] Update the estimates of the parameters. 
\item[Step 5] Repeat Step 3 and Step 4 until convergence. 
\end{description}

\subsection{Selection of tuning parameters}
\label{SelectionTuning}

We have four tuning parameters. 
To reduce the computational cost, the two tuning parameters $w$ and $\xi$ are fixed in advance. 
The tuning parameter $w$ is set to a small value, because the regression loss is more important than the PCA loss. 
The tuning parameter $\xi$ is also set to a small value, because the sparse regularization is more important than the ridge regularization. 
The latter idea is often used in the elastic net. 
The remaining parameters $\lambda_\beta$ and $\lambda_\gamma$ are automatically selected by cross-validation.


Let us divide the original dataset into $K$ datasets  $({\bm y}^{(1)}, X^{(1)}), \ldots, ({\bm y}^{(K)}, X^{(K)})$. 
The $K$-fold cross-validation criterion for aSPCR-glm is given by
\begin{eqnarray*}
{\rm CV_{glm}} = - \frac{1}{K} \sum_{k=1}^K  \sum_{i \in C_k} \left\{ \frac{y_i (\hat\gamma_0^{(-k)} + \hat{\bm \gamma}^{(-k)T} \hat{B}^{(-k)T} {\bm x}_i) - u(\hat\gamma_0^{(-k)} + \hat{\bm \gamma}^{(-k)T} \hat{B}^{(-k)T} {\bm x}_i) }{\hat\phi^{(-k)}} + v(y_i, \hat\phi^{(-k)}) \right\},
\end{eqnarray*}
where $C_k \ (k=1,\ldots,K)$ is the set of indexes for the divided dataset $({\bm y}^{(k)}, X^{(k)})$, and $\hat{\gamma}_0^{(-k)}, \hat{B}^{(-k)}, \hat{\bm \gamma}^{(-k)}, \hat\phi^{(-k)}$ are the estimates computed with the data removing the $k$-th part. 
The $K$-fold cross-validation criterion for the multiclass-logistic model is given by
\begin{eqnarray*}
{\rm CV_{multi}} = - \frac{1}{K} \sum_{k=1}^K \sum_{g=1}^G \sum_{i \in C_k}  [ y_{ig} (\hat\gamma_{0g}^{(-k)} + \hat{\bm \gamma}_g^{(-k)T} \hat{B}^{(-k)T} {\bm x}_i) - \log \{ 1 + \exp(\hat\gamma_{0g}^{(-k)} + \hat{\bm \gamma}_g^{(-k)T} \hat{B}^{(-k)T} {\bm x}_i) \} ],
\end{eqnarray*}
where $\hat{\gamma}_{0g}^{(-k)}, \hat{B}^{(-k)},$ and $\hat{\bm \gamma}_g^{(-k)}$ are the estimates computed with the data removing the $k$-th part. 
We employed $K=5$ in our simulation. 
The candidates of the regularization parameters $\lambda_{\beta}$ and $\lambda_{\gamma}$ were determined according to the function \texttt{glmnet} in \texttt{R}. 




\section{Illustrative example}
\label{IllustrativeExample}

We generated a dataset $\{ (y_i, \bm x_i); i=1,\ldots,200 \}$ for a binary response variable and 10-dimensional explanatory variables. 
The explanatory variables were given by ${\bm x}_i = P( {\bm u}_i^T, {\bm v}_i^T )^T$, where 
\begin{eqnarray*}
\bm u_i \sim \frac{1}{4} \sum_{j=1}^4 N_2 ( {\bm a}_j, 0.5^2 I_2 ), \quad \bm v_i \sim N_8 ( {\bm 0}, \Sigma ), \quad  P = {\rm block diag} (Q^T, I_{8}). 
\end{eqnarray*}
Here, ${\bm a}_1 = (2,2)^T, {\bm a}_2 = (-2,2)^T, {\bm a}_3 = (-2,-2)^T, {\bm a}_4 = (2,-2)^T$, $(\Sigma)_{ij} = 0.8^{| i-j |}$, and $Q$ is a 2 $\times$ 2 matrix whose $i$-th column is the eigenvector  corresponding to the $i$-th largest eigenvalue of ${\rm Var}( \bm u )$. 
Note that ${\bm u}_i$ presents four clusters which have four centers at the ${\bm a}_j$'s. 
We can easily show that ${\bm \nu}_1 = ( 0, 1, \underbrace{0, \ldots, 0}_8 )^T$ and ${\bm \nu}_2 = ( 1, \underbrace{0,  \ldots, 0}_9 )^T$ are the eigenvectors of ${\rm Var} (P {\bm u})$ and the third and fourth eigenvalues of ${\rm Var} ({\bm x})$, respectively. 
We also see $( {\bm \nu}_2^T {\bm x}_i,  {\bm \nu}_1^T {\bm x}_i)^T = {\bm u}_i$. 
The response variable $y_i$ was distributed according to the Bernoulli distribution with probability $\theta_i$ that satisfies $\log \{ \theta_i / (1-\theta_i) \} = {\bm x}_i^T {\bm \nu}_1 + {\bm x}_i^T {\bm \nu}_2$. 
This setting implies that the response is related to the principal components corresponding to the third and fourth eigenvalues of ${\rm Var} ({\bm x})$.

\begin{figure}[htbp]
\centering
\includegraphics[width=7cm,height=7cm]{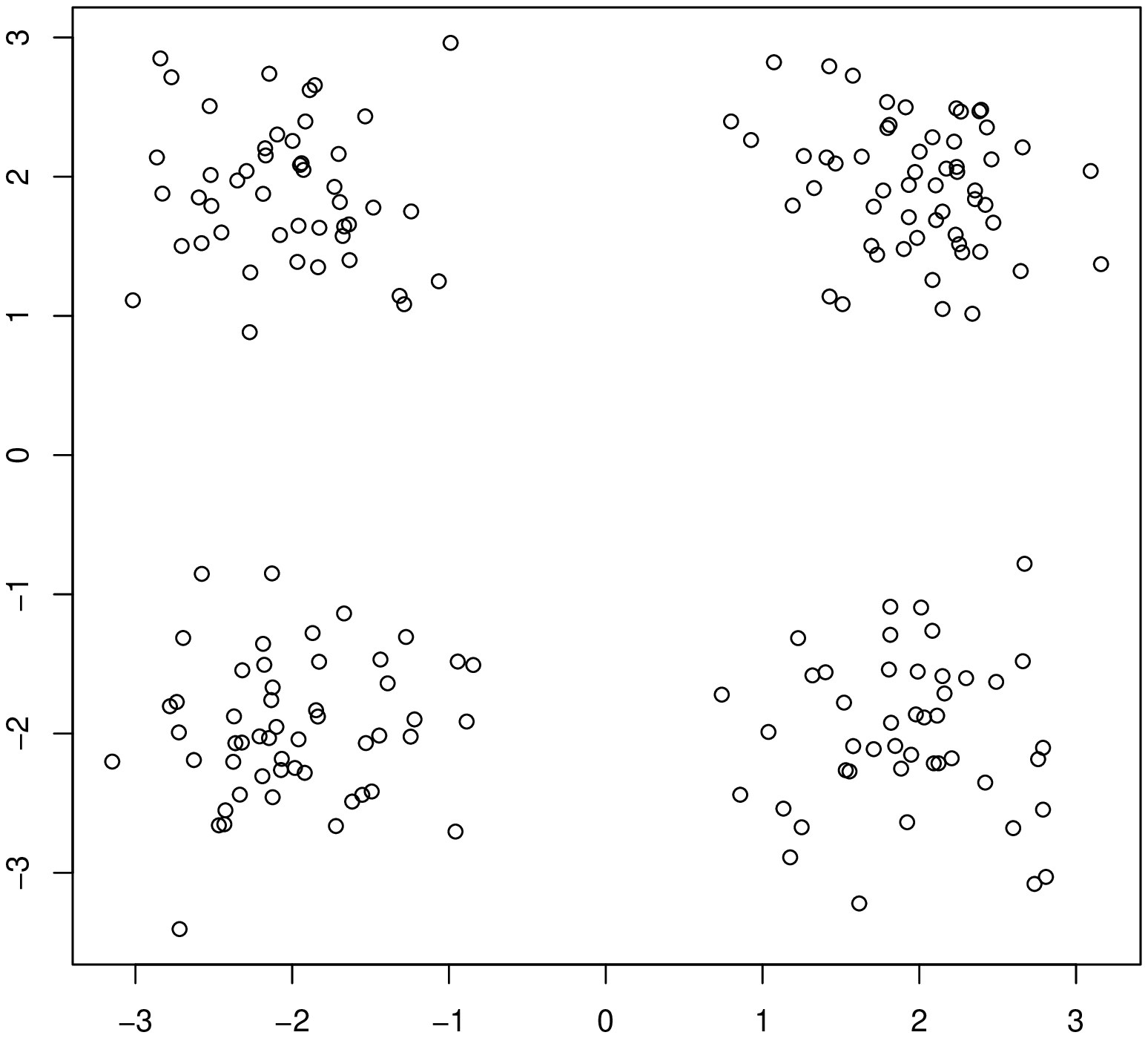}
\includegraphics[width=7cm,height=7cm]{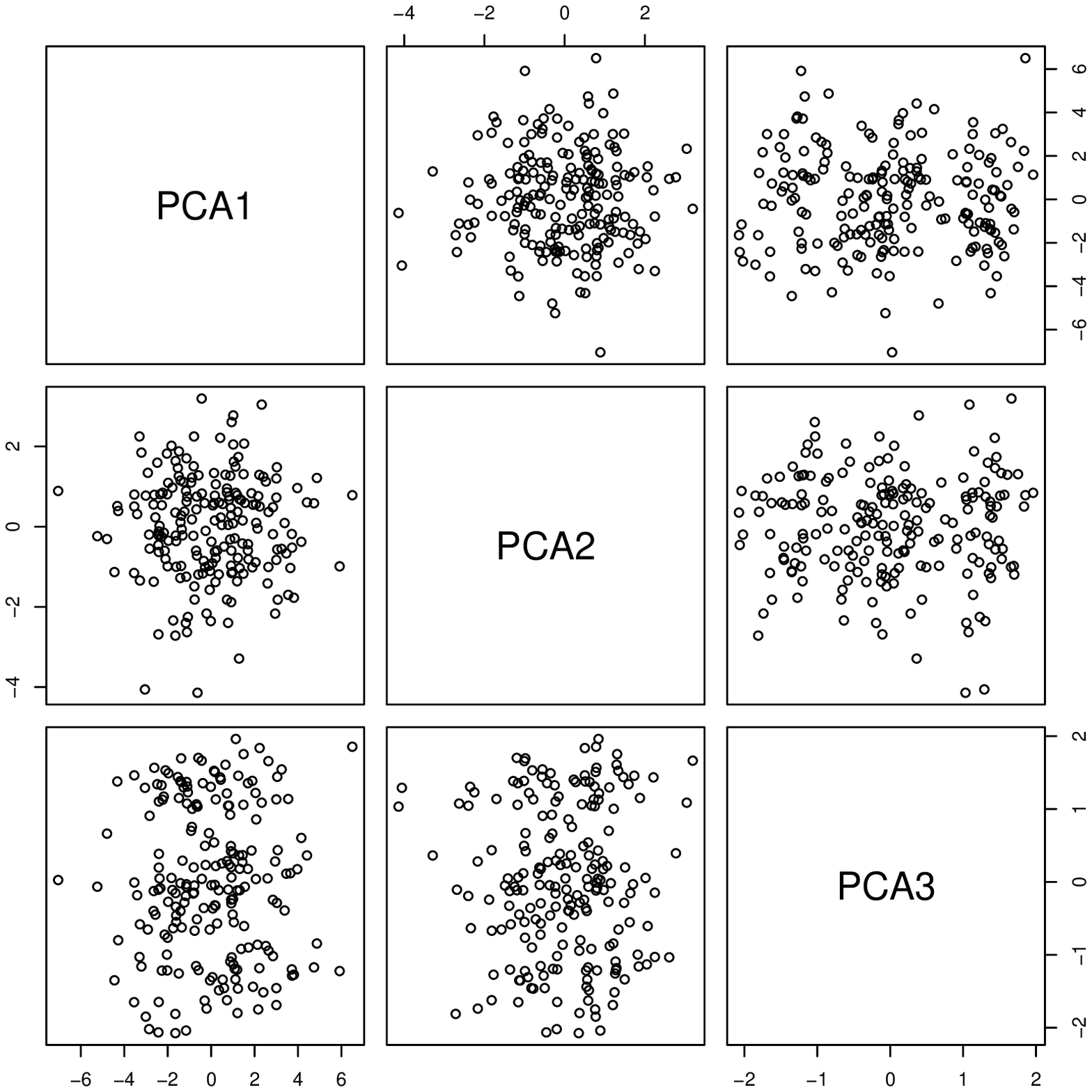}\\ [-0.5cm] 
\hspace{0.1cm} (a) \hspace{6.3cm} (b) \\[4.5mm] 
\includegraphics[width=7cm,height=7cm]{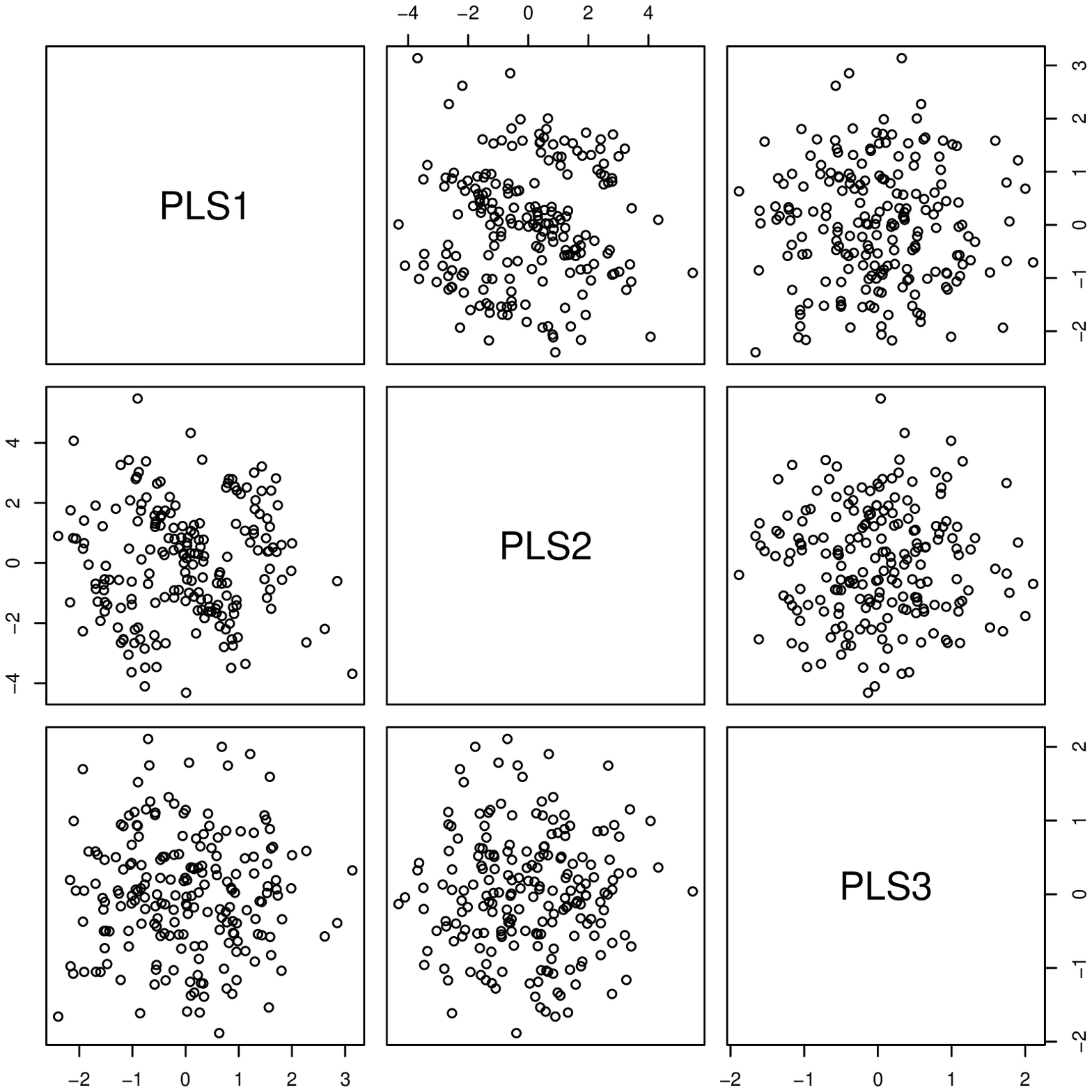}
\includegraphics[width=7cm,height=7cm]{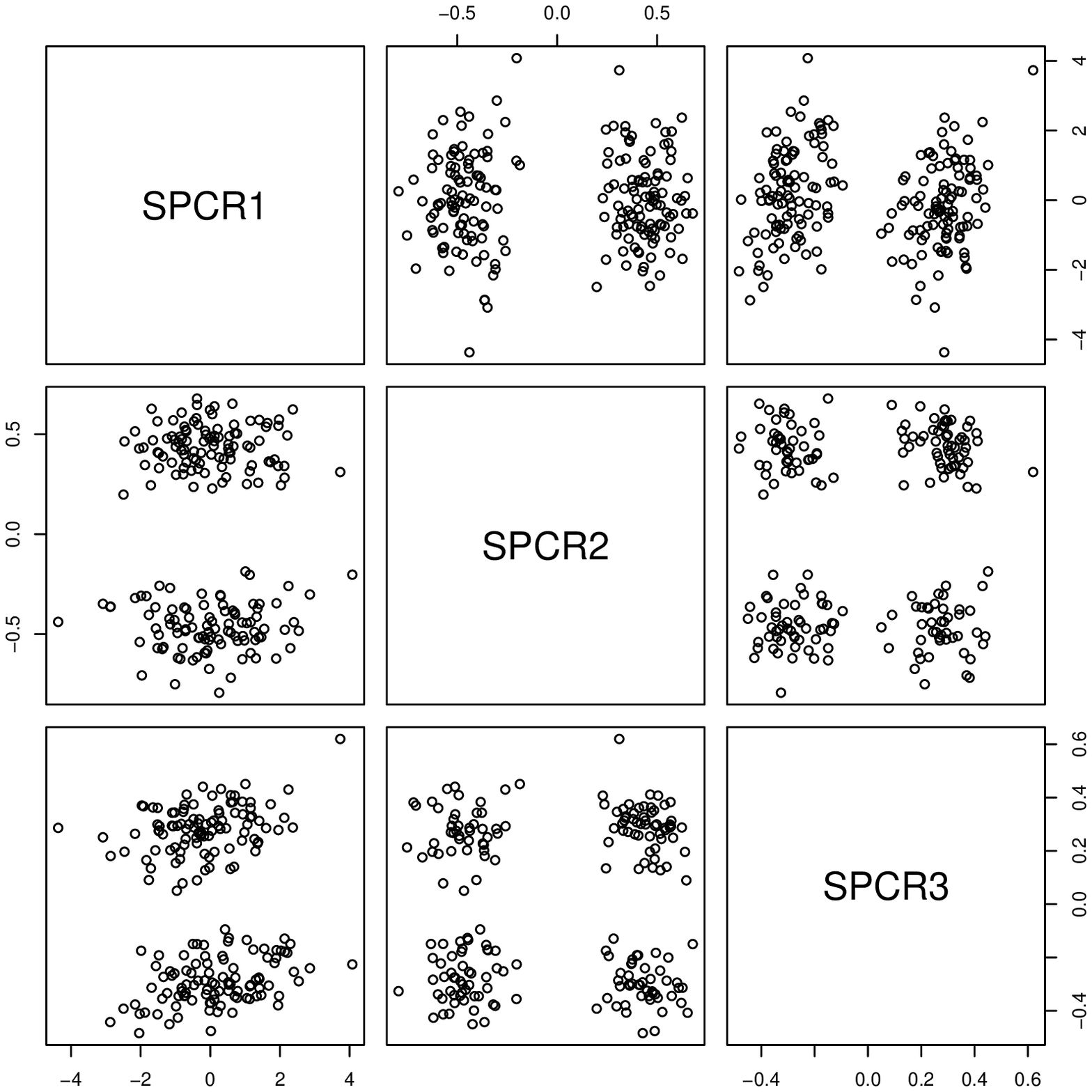}\\ [-0.5cm] 
\hspace{0.1cm} (c) \hspace{6.3cm} (d) \\[4.5mm] 
\caption{Scatter plots of principal and PLS components. 
(a) True structure of the principal components. 
(b) PCA. 
(c) PLS-GLR. 
(d) SPCR-glm. }
\label{ToyExample}
\end{figure}

We applied SPCR-glm with $k=3$ to the dataset, and then conducted PCA and PLS-GLR as described in Section \ref{RelatedWork}. 
The tuning parameters in SPCR-glm were set to $w=0.01, \xi=0.001, \lambda_\beta=3$, and $\lambda_\gamma=0.1$. 
Figure \ref{ToyExample} shows the true scatter plot of the $( {\bm \nu}_2^T {\bm x}_i,  {\bm \nu}_1^T {\bm x}_i)^T = {\bm u}_i$'s in (a) and the scatter PC plots for PCA in (b), for PLS-GLR in (c), and  for our proposed method in (d). 
We observe that (b) and (c) fail to capture the {true structure}, while (d) {succeeds} in finding the true structure {by  the second and third principal components. }

\section{Simulation studies}
\label{NumericalStudy}
To investigate the performances of our proposed method, Monte Carlo simulations were conducted. 
We used four models in this study: two for binary data and two for count data.

In the first model, we considered a 20-dimensional covariate vector ${\bm x} =(x_1,\ldots,x_{20})^T$ following multivariate normal distribution $N({\bm 0}_{20}, \Sigma_2)$, and generated the response $y$ \textcolor{black}{by}
\begin{eqnarray*}
y_i \sim B (1, p_i), \quad \log \left( \frac{p_i}{1-p_i} \right) =  4 {\bm x}_i^T {\bm \xi}^*, \quad i=1,\ldots,n.
\end{eqnarray*}
\textcolor{black}{
We used $\Sigma_2 = {\rm block diag} (\Sigma_{2}^*, I_{11})$ and ${\bm \xi}^* = ({\bm \nu}_1^*,0,\ldots,0)^T$, where $\left( \Sigma_{2}^* \right)_{ij} = 0.9^{|i-j|} \ (i,j=1,\ldots,9)$ and  ${\bm \nu}_1^*=(-1,0,1,1,0,-1,-1,0,1)$ is a sparse approximation of the fourth eigenvector of $\Sigma_{2}^*$.
}

In the second model, we considered a 30-dimensional covariate vector ${\bm x} =(x_1,\ldots,x_{30})^T$ following multivariate normal distribution $N({\bm 0}_{30}, \Sigma_3)$, and generated the response $y$ \textcolor{black}{by}
\begin{eqnarray*}
y_i \sim B (1, p_i), \quad \log \left( \frac{p_i}{1-p_i} \right) = 2 {\bm x}_i^T {\bm \xi}^{*}_1 + 2 {\bm x}_i^T {\bm \xi}^{*}_2, \quad i=1,\ldots,n.
\end{eqnarray*}
We used $\Sigma_3={\rm block diag} (\Sigma_{2}^*, \Sigma_{3}^*, I_{15})$ with $\left( \Sigma_{3}^* \right)_{ij} = 0.9^{|i-j|} \ (i,j=1,\ldots,6)$, ${\bm \xi}^{*}_1 = ({\bm \nu}^{*}_{1}, 0,\ldots,0)^T$, and ${\bm \xi}_2^*=(\underbrace{0,\ldots,0}_9,{\bm \nu}_{2}^*,\underbrace{0,\ldots,0}_{15})^T$, where {${\bm \nu}^*_{2} = (1, 0, -1, -1, 0, 1)$}  is a sparse approximation of the third eigenvector of $\Sigma_{3}^*$.

In the third model, we considered a 20-dimensional covariate vector ${\bm x} =(x_1,\ldots,x_{20})^T$ following multivariate normal distribution $N({\bm 0}_{20}, \Sigma_2)$, and generated the response $y$ \textcolor{black}{by}
\begin{eqnarray*}
y_i \sim Poi (\lambda_i), \quad \log \left( \lambda_i \right) = 0.8{\bm x}_i^T {\bm \xi}^*, \quad i=1,\ldots,n.
\end{eqnarray*}

In the fourth model, we considered a 30-dimensional covariate vector ${\bm x} =(x_1,\ldots,x_{30})^T$ following multivariate normal distribution $N({\bm 0}_{30}, \Sigma_3)$, and generated the response $y$ \textcolor{black}{by}
\begin{eqnarray*}
y_i \sim Poi (\lambda_i), \quad \log \left( \lambda_i \right) = 0.5 {\bm x}_i^T {\bm \xi}^{*}_1 + 0.5 {\bm x}_i^T {\bm \xi}^{*}_2, \quad i=1,\ldots,n.
\end{eqnarray*}

\begin{table}[t]
\begin{center}
\small
\caption{Mean (standard deviation) values of the EL for Cases 1 and 2.
The bold values correspond to the smallest means. }
\vspace{5mm}
\begin{tabular}{@{\extracolsep{-5.5pt}}ccccccccc} \hline
Case & $n$ & $k$ & aSPCR-Log(0.1) & aSPCR-Log(0.5) & aSPCR-Log(1) & SPCR-Log & PCR & PLS \\
\hline
1 & 200 & 1 & 0.344 & \textbf{0.328} & 0.421 & 0.347 & 0.697 & 0.666 \\
& &  & (0.100) & (0.092) & (0.184) & (0.098) & (0.005) & (0.035) \\
   &        & 5 & 0.354 & \textbf{0.316} & 0.688 & 0.365 & 0.701 & 0.366 \\
& &  & (0.119) & (0.067) & (0.038) & (0.127) & (0.009) & (0.048) \\
\hline
   & 400 & 1 & 0.301 & 0.287 & \textbf{0.285} & 0.299 & 0.695 & 0.633 \\
& &  & (0.071) & (0.043) & (0.043) & (0.059) & (0.002) & (0.046) \\
   &        & 5 & 0.375 & \textbf{0.287} & 0.678 & 0.388 & 0.696 & 0.301 \\
& &  & (0.165) & (0.043) & (0.066) & (0.172) & (0.004) & (0.025) \\
\hline
2 & 200 & 2 & 0.455 & \textbf{0.449} & 0.692 & 0.460 & 0.698 & 0.675 \\
& &  & (0.046) & (0.041) & (0.021) & (0.047) & (0.008) & (0.040) \\
   &        & 5 & \textbf{0.455} & 0.468 & 0.695 & 0.457 & 0.703 & 0.509 \\
& &  & (0.050) & (0.081) & (0.003) & (0.046) & (0.011) & (0.074) \\
\hline
   & 400 & 2 & 0.411 & \textbf{0.401} & 0.691 & 0.410 & 0.693 & 0.614 \\
& &  & (0.035) & (0.019) & (0.027) & (0.020) & (0.005) & (0.034) \\
   &        & 5 & 0.413 & \textbf{0.401} & 0.693 & 0.415 & 0.695 & 0.427 \\
& &  & (0.043) & (0.018) & (0.003) & (0.043) & (0.006) & (0.028) \\
\hline
\end{tabular}
\label{Table_KL_12}
\end{center}
\end{table}

\begin{table}[htbp]
\begin{center}
\small
\caption{Mean (standard deviation) values of the EL for Cases 3 and 4.
The bold values correspond to the smallest means. }
\vspace{5mm}
\begin{tabular}{@{\extracolsep{-5.5pt}}ccccccccc} \hline
Case & $n$ & $k$ & aSPCR-Poi(0.1) & aSPCR-Poi(0.5) & aSPCR-Poi(1) & SPCR-Poi & PCR & PLS \\
\hline
3 & 200 & 1 & 1.386 & 1.378 & \textbf{1.375} & 1.395 & 1.915 & 1.875 \\
 &  &  & (0.060) & (0.079) & (0.082) & (0.079) & (0.071) & (0.092) \\
 &  & 5 & 1.382 & \textbf{1.368} & 1.382 & 1.389 & 1.932 & 1.392 \\
 &  &  & (0.042) & (0.042) & (0.075) & (0.071) & (0.075) & (0.039) \\
 \hline
 & 400 & 1 & 1.351 & \textbf{1.345} & 1.347 & 1.353 & 1.886 & 1.780 \\
 &  &  & (0.086) & (0.087) & (0.086) & (0.086) & (0.071) & (0.105) \\
 &  & 5 & 1.345 & \textbf{1.336} & 1.346 & 1.347 & 1.891 & 1.347 \\
 &  &  & (0.067) & (0.041) & (0.066) & (0.067) & (0.070) & (0.036) \\
 \hline
4 & 200 & 2 & 1.403 & \textbf{1.394} & 1.510 & 1.406 & 1.658 & 1.625 \\
 &  &  & (0.045) & (0.046) & (0.113) & (0.045) & (0.043) & (0.055) \\
 &  & 5 & 1.404 & \textbf{1.399} & 1.552 & 1.407 & 1.665 & 1.423 \\
 &  &  & (0.050) & (0.052) & (0.115) & (0.049) & (0.045) & (0.040) \\
 \hline
 & 400 & 2 & 1.360 & \textbf{1.357} & 1.439 & 1.361 & 1.655 & 1.566 \\
 &  &  & (0.039) & (0.037) & (0.099) & (0.039) & (0.047) & (0.053) \\
 &  & 5 & 1.360 & \textbf{1.359} & 1.477 & 1.362 & 1.659 & 1.368 \\
 &  &  & (0.036) & (0.036) & (0.122) & (0.037) & (0.047) & (0.033) \\
\hline
\end{tabular}
\label{Table_KL_34}
\end{center}
\end{table}

\begin{table}[htbp]
\begin{center}
\small
\caption{Mean (standard deviation) values of TPR and TNR for Cases 1 and 2.
The bold values correspond to the largest means. }
\vspace{5mm}
\begin{tabular}{@{\extracolsep{-5.5pt}}cccccccc} \hline
Case & $n$ & $k$ &  & aSPCR-Log(0.1) & aSPCR-Log(0.5) & aSPCR-Log(1) & SPCR-Log \\
\hline
1 & 200 & 1 & TPR & 0.930 & \textbf{0.956} & 0.736 & 0.933 \\
 &  &  &  & (0.256) & (0.176) & (0.394) & (0.246) \\
 &  &  & TNR & 0.267 & 0.580 & \textbf{0.920} & 0.190 \\
 &  &  &  & (0.228) & (0.201) & (0.086) & (0.240) \\
 &  & 5 & TPR & 0.928 & \textbf{0.980} & 0.048 & 0.915 \\
 &  &  &  & (0.210) & (0.114) & (0.123) & (0.232) \\
 &  &  & TNR & 0.349 & 0.585 & \textbf{1} & 0.270 \\
 &  &  &  & (0.260) & (0.187) & (0) & (0.284) \\
 \hline
 & 400 & 1 & TPR & 0.973 & \textbf{0.993} & 0.991 & 0.983 \\
 &  &  &  & (0.154) & (0.066) & (0.083) & (0.119) \\
 &  &  & TNR & 0.349 & 0.585 & \textbf{1} & 0.270 \\
 &  &  &  & (0.260) & (0.187) & (0) & (0.284) \\
 &  & 5 & TPR & 0.876 & \textbf{0.991} & 0.088 & 0.863 \\
 &  &  &  & (0.236) & (0.083) & (0.184) & (0.244) \\
 &  &  & TNR & 0.349 & 0.585 & \textbf{1} & 0.270 \\
 &  &  &  & (0.260) & (0.187) & (0) & (0.284) \\
 \hline
2 & 200 & 2 & TPR & 0.980 & \textbf{0.991} & 0.018 & 0.982 \\
 &  &  &  & (0.112) & (0.032) & (0.103) & (0.112) \\
 &  &  & TNR & 0.278 & 0.474 & \textbf{1} & 0.190 \\
 &  &  &  & (0.167) & (0.154) & (0) & (0.154) \\
 &  & 5 & TPR & 0.977 & 0.906 & 0.003 & \textbf{0.984} \\
 &  &  &  & (0.107) & (0.243) & (0.017) & (0.095) \\
 &  &  & TNR & 0.300 & 0.519 & \textbf{1} & 0.197 \\
 &  &  &  & (0.178) & (0.221) & (0) & (0.155) \\
 \hline
 & 400 & 2 & TPR & 0.990 & \textbf{1} & 0.012 & \textbf{1} \\
 &  &  &  & (0.100) & (0) & (0.100) & (0) \\
 &  &  & TNR & 0.300 & 0.519 & \textbf{1} & 0.197 \\
 &  &  &  & (0.178) & (0.221) & (0) & (0.155) \\
 &  & 5 & TPR & 0.987 & \textbf{1} & 0.004 & 0.982 \\
 &  &  &  & (0.093) & (0) & (0.019) & (0.127) \\
 &  &  & TNR & 0.331 & 0.612 & \textbf{1} & 0.218 \\
 &  &  &  & (0.149) & (0.133) & (0) & (0.143) \\
\hline
\end{tabular}
\label{Table_TPRTNR_12}
\end{center}
\end{table}

\begin{table}[htbp]
\begin{center}
\small
\caption{Mean (standard deviation) values of TPR and TNR for Cases 3 and 4.
The bold values correspond to the largest means. }
\vspace{5mm}
\begin{tabular}{@{\extracolsep{-5.5pt}}cccccccc} \hline
Case & $n$ & $k$ &  & aSPCR-Poi(0.1) & aSPCR-Poi(0.5) & aSPCR-Poi(1) & aSPCR-Poi \\
\hline
3 & 200 & 1 & TPR & \textbf{0.993} & 0.981 & 0.973 & 0.980 \\
 &  &  &  & (0.066) & (0.118) & (0.143) & (0.140) \\
 &  &  & TNR & 0.247 & 0.607 & \textbf{0.921} & 0.150 \\
 &  &  &  & (0.160) & (0.198) & (0.115) & (0.161) \\
 &  & 5 & TPR & \textbf{0.996} & 0.995 & 0.980 & 0.990 \\
 &  &  &  & (0.033) & (0.037) & (0.106) & (0.100) \\
 &  &  & TNR & 0.250 & 0.659 & \textbf{0.964} & 0.165 \\
 &  &  &  & (0.152) & (0.183) & (0.068) & (0.140) \\
 \hline
 & 400 & 1 & TPR & \textbf{0.980} & \textbf{0.980} & \textbf{0.980} & \textbf{0.980} \\
 &  &  &  & (0.140) & (0.140) & (0.140) & (0.140) \\
 &  &  & TNR & 0.287 & 0.737 & \textbf{0.987} & 0.182 \\
 &  &  &  & (0.183) & (0.183) & (0.055) & (0.165) \\
 &  & 5 & TPR & 0.990 & \textbf{0.996} & 0.990 & 0.990 \\
 &  &  &  & (0.100) & (0.033) & (0.100) & (0.100) \\
 &  &  & TNR & 0.305 & 0.800 & \textbf{0.994} & 0.180 \\
 &  &  &  & (0.175) & (0.166) & (0.021) & (0.144) \\
 \hline
4 & 200 & 2 & TPR & 0.971 & 0.953 & 0.526 & \textbf{0.976} \\
 &  &  &  & (0.143) & (0.145) & (0.396) & (0.142) \\
 &  &  & TNR & 0.270 & 0.659 & \textbf{0.974} & 0.172 \\
 &  &  &  & (0.151) & (0.150) & (0.050) & (0.149) \\
 &  & 5 & TPR & 0.964 & 0.938 & 0.366 & \textbf{0.969} \\
 &  &  &  & (0.156) & (0.159) & (0.397) & (0.150) \\
 &  &  & TNR & 0.296 & 0.691 & \textbf{0.982} & 0.189 \\
 &  &  &  & (0.161) & (0.151) & (0.042) & (0.170) \\
 \hline
 & 400 & 2 & TPR & \textbf{0.990} & 0.986 & 0.766 & \textbf{0.990} \\
 &  &  &  & (0.100) & (0.073) & (0.336) & (0.100) \\
 &  &  & TNR & 0.291 & 0.749 & \textbf{0.992} & 0.183 \\
 &  &  &  & (0.145) & (0.147) & (0.019) & (0.129) \\
 &  & 5 & TPR & \textbf{0.993} & 0.987 & 0.635 & 0.992 \\
 &  &  &  & (0.070) & (0.082) & (0.411) & (0.080) \\
 &  &  & TNR & 0.304 & 0.754 & \textbf{0.992} & 0.192 \\
 &  &  &  & (0.149) & (0.157) & (0.022) & (0.135) \\
\hline
\end{tabular}
\label{Table_TPRTNR_34}
\end{center}
\end{table}


The sample size was set to $n=200, 400$. 
For Cases 1 and 2, we used SPCR-glm for binary data (SPCR-Log) and aSPCR-glm for binary data with $q=0.1, 0.5, 1$ (aSPCR-Log($q$)). 
For Cases 3 and 4, we used SPCR-glm for count data (SPCR-Poi) and aSPCR-glm for count data with $q=0.1, 0.5, 1$ (aSPCR-Poi($q$)). 
The proposed methods were fitted to the simulated data with one or five components $(k=1,5)$ for Cases 1 and 3, and two or five components $(k=2,5)$ for Cases 2 and 4. 
The regularization parameters $\lambda_{\beta}$ and $\lambda_{\gamma}$ were selected by five-fold cross-validation as described in Section \ref{SelectionTuning}. 
The tuning parameters $w$ and $\zeta$ were set to 0.01 and 0.001, respectively. 
Our proposed methods were compared with PLS-GLR and PCR. 
The performance was evaluated in terms of the value of the negative expected log-likelihood function $-E \left[  \log f( y | {\bm x} ; \hat{\bm \theta} )  \right]$ (EL). 
The simulation was conducted 100 times. 
EL was estimated by 1,000 random samples.

Tables \ref{Table_KL_12} and \ref{Table_KL_34} list the means and standard deviations of ELs for Cases 1 and 2, and these two results show similar tendencies. 
PCR was the worst in all cases. 
Our proposed methods outperform other methods when $k=1, 2$, and were competitive with PLS-GLR when $k=5$. 
The proposed method with $q=0.5$ was superior to other methods in almost all cases. 
The smallest ELs were provided by the proposed method with $q=1$ in Case 1 for $n=400$ and $k=1$ and in Case 3 for $n=200$ and $k=1$. 
The performance of aSPCR-Log(0.1) or aSPCR-Poi(0.1) was similar to that of aSPCR-Log or aSPCR-Poi, respectively.


%

We also computed the true positive rate (TPR) and the true negative rate (TNR) for aSPCR-Log($q$), SPCR-Log, aSPCR-Poi($q$), and SPCR-Poi, which are defined by
\begin{eqnarray*}
\mathrm{TPR}&=&
\frac{1}{100}
\sum_{k=1}^{100}
\frac{\left|\left\{ 
j:\hat{\zeta}^{(k)}_{j}\neq0~\wedge~\zeta^{\ast}_{j}\neq 0
\right\}\right|}
{\left|\left\{
j:\zeta^{\ast}_{j}\neq 0 
\right\}\right|}, \\
\mathrm{TNR}&=&
\frac{1}{100}
\sum_{k=1}^{100}
\frac{\left|\left\{
j:\hat{\zeta}^{(k)}_{j}=0~\wedge~\zeta^{\ast}_{j}=0
\right\}\right|}
{\left|\left\{
j:\zeta^{\ast}_{j}= 0 
\right\}\right|}, 
\end{eqnarray*}
where $\hat{\zeta}^{(k)}_{j}$ is the estimated $j$-th coefficient for the $k$-th simulation, and $|\{\ast\}|$ is the number of elements included in a set $\{\ast\}$. 
Tables \ref{Table_TPRTNR_12} and \ref{Table_TPRTNR_34} list the means and standard deviations of TPR and TNR, and present similar results. 
In Table \ref{Table_TPRTNR_12}, many methods provide higher ratios of TPRs except for aSPCR-Log(1), while aSPCR-Log(1) provides higher ratios of TNRs. 
The results for Cases 3 and 4 show that the TPRs are higher in almost all situations, but the TNRs of aSPCR-Poi(0.1) and aSPCR-Poi are too much lower.




\section{Applications}
\label{Applications}
In this section, two real data analyses are illustrated. 
With the proposed method, we observe more easily interpretable PC scores and clearer classification on PC plots than using the usual PCA.

\subsection{Doctor visits data}
\label{DoctorVisits}
We applied SPCR-glm to the doctor visits data in Cameron and Trivedi (1986). 
This dataset consists of 5,190 observations originating from the Australian Health Survey and contains information on the number of consultations with a doctor or specialist and on 11 variables: (1) Gender,  (2) Age, (3) Income, (4) Illness, (5) Reduced, (6) Health, (7) Private, (8) Freepoor, (9) Freerepeat, (10) Nchronic, and (11) Lchronic. 
The dataset is available in the package {\ttfamily AER} for {\ttfamily R}.

To model a relationship between the number of consultations, which is count data, and the 11 variables, we utilized SPCR-glm with $k=5$. 
We compared SPCR-glm with PCR and PLS-GLR. 
The tuning parameters in SPCR-glm were set as $w=0.1, \xi=0.001, \lambda_\gamma=0, \lambda_\beta=10$. 
The reason for using $\lambda_\gamma=0$ was that we were not trying to select the number of principal components automatically.

\begin{figure}[htbp]
\centering
\includegraphics[width=10cm,height=8cm]{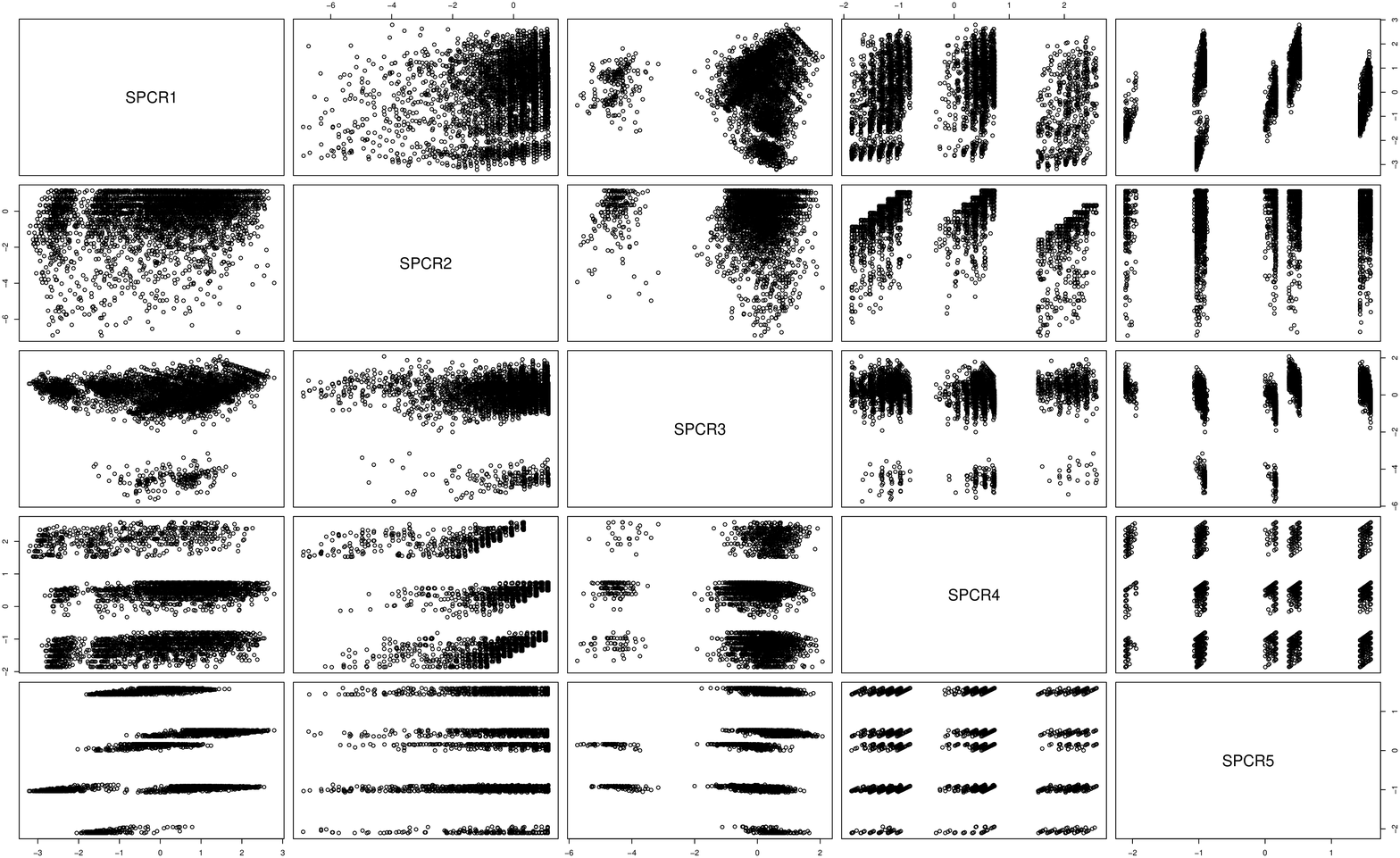}
\caption{Scatter plots of principal components given by SPCR-glm for the doctor visits data. }
\label{fig:RealDataPoi}
\end{figure}

\begin{table}[htbp]
\begin{center}
\small
\caption{Estimates of $B$ for the doctor visits data. }
\vspace{5mm}
\begin{tabular}{@{\extracolsep{-5.5pt}}lccccc} \hline
variable & PC1 & PC2 & PC3 & PC4 & PC5 \\
\hline
Gender & --0.535 & --0.011 & 0.082 & 0 & 0.535 \\
Age & --0.451 & 0 & --0.322 & --0.090 & --0.062 \\
Income & 0.497 & 0 & --0.351 & 0 & 0 \\
Illness & --0.047 & 0.530 & 0 & --0.226 & 0 \\
Reduced & 0.019 & 0.688 & --0.085 & 0 & 0 \\
Health & 0.061 & 0.416 & 0.212 & --0.002 & 0 \\
Private & 0.084 & 0.008 & --0.195 & 0 & 0.710 \\
Freepoor & 0 & 0 & 0.779 & 0 & 0 \\
Freerepeat & --0.459 & 0 & --0.152 & 0 & --0.422 \\
Nchronic & --0.034 & 0.043 & --0.032 & --0.751 & 0 \\
Lchronic & --0.131 & 0.259 & --0.089 & 0.594 & 0 \\
\hline
\end{tabular}
\label{table:RealDataPoi}
\end{center}
\end{table}

Figure \ref{fig:RealDataPoi} gives the PC scatter plots for SPCR-glm. 
Some clusters are observed with the inclusion of the third through fifth principal components, PC3, PC4, and PC5. 
This may imply that the dataset has several natural clusters. 
This finding is not seen in the PC scatter plots for PCR and PLS-GLR (Appendix B). 
Furthermore, we performed the five-fold cross validation for SPCR-glm, PCR, and PLS-GLR, to compare the prediction performance. 
The validation values were $0.652$, $0.662$, and $0.651$, respectively.

The estimates $\hat{\gamma}_0$ and $\hat{\bm \gamma}$ for SPCR-glm were given by
\begin{eqnarray*}
\hat{\gamma}_0 = -1.484, \quad \hat{\bm \gamma} = (-0.106, 0.433, -0.124, -0.087, 0.065)^T,
\end{eqnarray*}
and the estimate of the loading matrix $B$ is shown in Table \ref{table:RealDataPoi}. 
We can interpret some of the principal component loadings as follows. 
PC2 represents an index of health state, because it provides larger values for Illness, Reduced, and Health factors, where the first and second factors are a clinical history over the past two weeks and the third factor is a general health index. 
PC3 shows whether it is easy to visit the hospital, because it provides larger values for the Freepoor factor and smaller values for the age factor, where Freepoor indicates whether or not the individual has a free government health insurance due to low income. 
PC4 represents an overall index for chronic disease, because it provides larger values for the Lchronic factor and smaller values for the Illness and Nchronic factors, where Nchronic is a chronic condition not limiting activity, while Lchronic is that limiting activity. 
Meanwhile, it is difficult to interpret the principal component loadings for PCR and PLS-GLR (Appendix B).


\subsection{{Mouse consomic strain data}}
\label{Genomic}

Takada \textit{et al.} (2008) provided the dataset on mouse inter-subspecific consomic strains. 
The consomic strain (CS) was made from the standard strain C57BL/6 (B6) by replacing a chromosome by the corresponding chromosome in the strain MSM/Ms (MSM). 
The phenotypes of B6 are different from those of MSM. 
If several phenotypes of the CS are similar to those of MSM, we can assume that the genetic factor corresponding to the phenotypes depends on the replaced chromosome. 
There were $G=30$ strains, including B6, MSM, and 28 CSs, with $p=36$ traits. 
Each {strain} had 7--16 animals.  
{Various characteristics of mice are given} in Takada \textit{et al.} (2008) and at the website \url{http://molossinus.lab.nig.ac.jp/phenotype/}. 
This dataset can be downloaded from the web server \url{ftp://molossinus.lab.nig.ac.jp/pub/phenotypedb/CONSOMIC_10W/}. 

\begin{figure}[htbp]
\begin{minipage}{.5\textwidth}
\centering
\includegraphics[width=7cm,height=7cm]{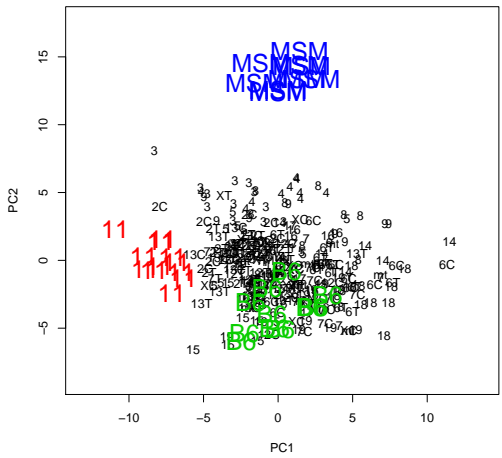} 
\subcaption{ {PC1 vs. PC2 for SPCR.}}
\label{fig:SPCR12}
\end{minipage}
\begin{minipage}{.5\textwidth}
\centering
\includegraphics[width=7cm,height=7cm]{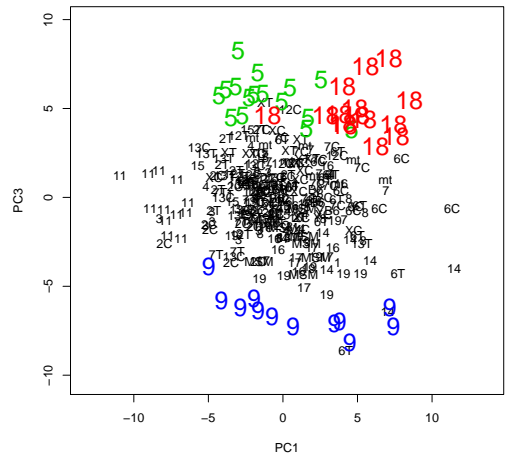} 
\subcaption{ {PC1 vs. PC3 for SPCR.}}
\label{fig:SPCR13}
\end{minipage}
\begin{minipage}{.5\textwidth}
\centering
\includegraphics[width=7cm,height=7cm]{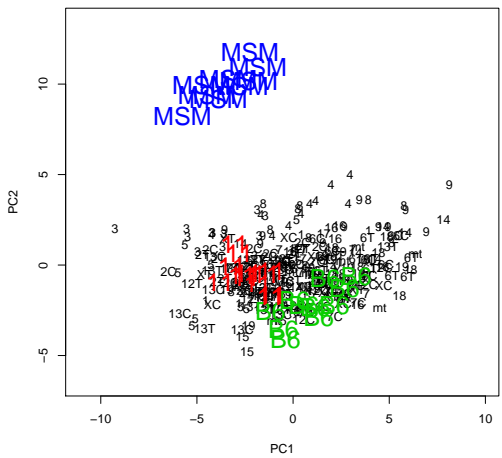} 
\subcaption{ {PC1 vs. PC2 for PCA.}}
\label{fig:PCA12}
\end{minipage}
\begin{minipage}{.5\textwidth}
\centering
\includegraphics[width=7cm,height=7cm]{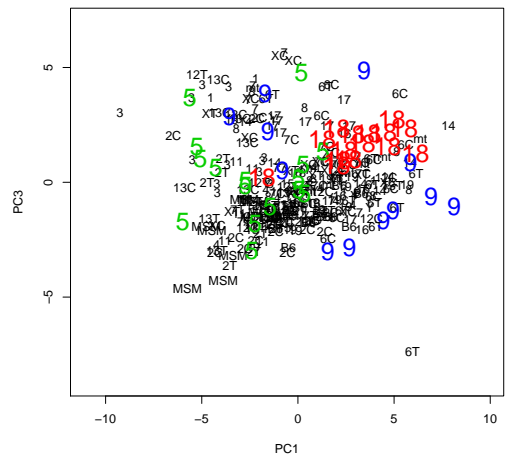} 
\subcaption{ {PC1 vs. PC3 for PCA.}}
\label{fig:PCA13}
\end{minipage}
\caption{Scatter plots of PC scores for the {mouse consomic strain data}.} 
\label{fig:PCplot_mouse}
\end{figure}

{
We analyzed the male data in this dataset by the usual PCA and by our method SPCR-glm with $k=3, w=0.01, \xi=0.001, \lambda_\beta=4,$ and $\lambda_\gamma=10$. 
There were very few missing observations with the ratio 0.39\%, which were imputed using the package {\ttfamily mice} for {\ttfamily R}. 
The resulting PC plots are shown in Figure~\ref{fig:PCplot_mouse} and the PC loadings and regression coefficients given by SPCR-glm are listed in Tables \ref{table:RealDataGenomic_B} and \ref{table:RealDataGenomic_gamma}.  
}


\begin{table}[htbp]
\begin{center}
\small
\renewcommand{\arraystretch}{0.8}
\caption{Estimates of $B$ for the {mouse consomic strain data}. }
\vspace{5mm}
\begin{tabular}{@{\extracolsep{-3.5pt}}lccc} \hline
variable & PC1 & PC2 & PC3 \\
\hline
BODY\_WEIGHT & 0 & 0 & 0 \\
BODY\_LENGTH & 0 & --0.223 & 0 \\
TAIL\_LENGTH & 0 & --0.050 & 0 \\
HEAD\_BODY\_LENGTH & 0 & 0 & 0 \\
TESTIS\_AVERAGE & --0.148 & 0 & --0.869 \\
SPLEEN & 0 & 0 & 0 \\
LIVER & 0 & 0 & 0 \\
KIDNEY\_AVERAGE & --0.150 & 0.851 & --0.541 \\
HEART & 0 & 0.225 & --0.343 \\
EPIDIDYMAL & 0 & 0.360 & --0.610 \\
PERIRENAL & 1.740 & 0 & 0 \\
MESENTRIC & 0 & 0.350 & 0 \\
INGUINAL & 0.633 & 0.139 & 0 \\
DORSAL\_WHITE\_FAT & 0 & 0 & 0 \\
DORSAL\_BROWN\_FAT & 0 & 0 & --0.429 \\
BMI & 0 & 0 & 0 \\
VISCERAL & 0 & 0 & 0 \\
SUBCUTANEOUS & 0.468 & 0 & 0 \\
TOTAL\_FAT\_PAT\_WEIGHT & 0.359 & 0 & 0 \\
LEAN\_WEIGHT & 0 & --2.661 & 0 \\
AI & 0 & 0 & 0 \\
IP & 0 & 0 & --0.014 \\
HDL & 0 & 0 & 0 \\
T\_CHOLESTEROL & 0 & --0.467 & --1.787 \\
NON\_HDL & 0 & 0 & --0.393 \\
TG & --0.066 & 0 & 0.540 \\
AMYL & 0 & 0 & --0.005 \\
ALB1 & 0 & 0 & 0.869 \\
ALP & --1.266 & 0 & 0 \\
ALT & 0 & 0 & --0.200 \\
TBIL & 0.042 & 0 & 0 \\
BUN & 0 & 0 & 0.009 \\
CALCIUM & 0.340 & 0 & 0 \\
TP & 0.153 & 0 & 0.387 \\
GLOB & 0 & 0 & --0.046 \\
POTASSIUM & 0 & 0 & 0 \\
\hline
\end{tabular}
\label{table:RealDataGenomic_B}
\end{center}
\end{table}

\begin{table}[htbp]
\begin{center}
\small
\renewcommand{\arraystretch}{0.9}
\caption{Estimates of $\bm \gamma$ for the {mouse consomic strain data}. }
\vspace{5mm}
\begin{tabular}{@{\extracolsep{-3.5pt}}lccc} \hline
strain & $\gamma_1$ & $\gamma_2$ & $\gamma_3$ \\
\hline
B6 & 0 & --0.471 & 0 \\
MSM & 0 & 1.164 & 0 \\
CS1 & 0 & 0 & 0 \\
CS2C & --0.282 & 0.149 & --0.142 \\
CS2T & --0.203 & 0 & 0 \\
CS3 & --0.228 & 0.663 & 0 \\
CS4 & 0 & 0.685 & 0 \\
CS5 & 0 & 0 & 1.151 \\
CS6C & 0.353 & 0 & 0 \\
CS6T & 0.304 & 0 & 0 \\
CS7 & 0.040 & 0 & 0 \\
CS7C & 0.153 & --0.166 & 0.168 \\
CS7T & --0.177 & 0 & --0.029 \\
CS8 & 0.089 & 0.338 & 0 \\
CS9 & 0 & 0.231 & --1.037 \\
CS11 & --1.054 & 0 & 0 \\
CS12C & 0 & 0 & 0.115 \\
CS12T & --0.141 & 0 & 0 \\
CS13C & --0.260 & 0 & 0 \\
CS13T & 0 & 0 & 0 \\
CS14 & 0.228 & 0 & --0.317 \\
CS15 & --0.163 & --0.470 & 0 \\
CS16 & 0.026 & 0 & --0.065 \\
CS17 & 0 & 0 & --0.357 \\
CS18 & 0.506 & --0.003 & 0.923 \\
CS19 & 0 & --0.429 & --0.552 \\
CSXC & 0 & --0.042 & 0 \\
CSXT & 0 & 0 & 0.350 \\
CSY & 0 & 0 & 0 \\
mt & 0.092 & 0 & 0.181 \\
\hline
\end{tabular}
\label{table:RealDataGenomic_gamma}
\end{center}
\end{table}

The first PC (PC1) given by SPCR-glm largely depended on the score 
\begin{eqnarray*}
1.74 \times \verb|PERIPENAL| - 1.27 \times \verb|ALP|,
\end{eqnarray*}
where \verb|PERIPENAL| is the value for the amount of fat around the kidneys and \verb|ALP| is the alkaline phosphatase value. 
B6 and MSM presented PC1 values near zero, which implies that PC1 is an index related to CSs. 
CS11 clearly presented PC2 values near zero and had a lower value for PC1 and was separate from other mice in Figure~\ref{fig:PCplot_mouse}\subref{fig:SPCR12}. 
In fact, CS11 had a small PERIPENAL and a large ALP, which appears as a small PC1 value. 
In Figure~\ref{fig:PCplot_mouse}\subref{fig:PCA12}, CS11 was not separate from other mice.

The second PC (PC2) given by SPCR-glm largely depended on the score 
\begin{eqnarray*}
0.85 \times \verb|KIDNEY_AVERAGE| - 2.66 \times \verb|LEAN_WEIGT|,
\end{eqnarray*}
where \verb|LEAN_WEIGT| is the weight after removing the white fat pad and  \verb|KIDNEY_AVERAGE| is the mean of two kidney (right and left) weights divided by \verb|LEAN_WEIGHT|. 
The two above traits were selected from among many traits related to the weight. 
For example, within a given CS, the \verb|BODY_WEIGHT| largely depended on the individual, so that its within-variance was large, whereas the \verb|LEAN_WEIGHT| did not, and so its within-variance was small. 
This was why \verb|LEAN_WEIGHT| was more favorable than other traits related to the weight. 
In Figure~\ref{fig:PCplot_mouse}\subref{fig:SPCR12}, PC2 clearly separates the basic strain MSM from other strains. 
MSM comprised small mice with large kidneys and the consomic strains were a more similar to B6 than to MSM because the consomic strains were based on the B6, which is explained by PC2.

The third PC (PC3) given {by SPCR-glm} largely depended on the score 
\begin{eqnarray*}
-0.87 \times \verb|TESTIS_AVERAGE| -1.79 \times \verb|T_CHOLESTEROL| + 0.87 \times \verb|ALB1|, 
\end{eqnarray*}
where \verb|TESTIS_AVERAGE| is the mean of two testis weights, \verb|T_CHOLESTEROL| is the cholesterol value, and \verb|ALB1| is the albumin value. 
In Figure~\ref{fig:PCplot_mouse}\subref{fig:SPCR13}, CS5, CS9, and CS18 were separate from other mice, but not in Figure~\ref{fig:PCplot_mouse}\subref{fig:PCA13}. 
In fact, CS5 and CS18 had a small \verb|TESTIS_AVERAGE| and a small \verb|T_CHOLESTEROL| and CS9 had a very large \verb|T_CHOLESTEROL|. 
However, we have not yet found a significant property on \verb|ALB1|. 
This may imply that \verb|ALB1| has a significant property which has not been found.

CS1, CS13T, and CSY had zero regression coefficients, as shown in Table \ref{table:RealDataGenomic_gamma}. 
This implies that they did not have any significant property on the three PCs. 
In fact, their CSs were known to have no significant property.

\section{Conclusion}
\label{Concluding}

We presented a one-stage procedure for PCR in the framework of generalized linear models with sparse regularization. 
We called this procedure SPCR-glm. 
We showed that SPCR-glm enables various types of response variables to be treated. 
An estimation algorithm for SPCR-glm was obtained based on the coordinate descent algorithm. 
Through numerical experiments, our proposed method was demonstrated to be superior to competing methods in terms of prediction accuracy, TPR, TNR, and interpretability of the principal component loadings.

\section*{Acknowledgement}
S. K. was supported by Grant-in-Aid for Young Scientist (B) (15K15947) and Grants-in-Aid for Scientific Research on Innovative Areas (16H06429 and 16H06430). 



\begin{thebibliography}{999}








\bibitem{Bastien2} Bastien, P., Vinzi, V. E. and Tenenhaus, M. (2005). PLS generalised linear regression. \textit{Computational Statistics \& Data Analysis}, \textbf{48}, 17--46.


\bibitem{Cameron} Cameron, A.C. and Trivedi, P.K. (1986). Econometric models based on count data: Comparisons and applications of some estimators and tests. \textit{Journal of Applied Econometrics}, \textbf{1}, 29--53.




\bibitem{Choi} Choi, J., Zou, H. and Oehlert, G. (2011). A penalized maximum likelihood approach to sparse factor analysis. \textit{Statistics and Its Interface},  \textbf{3}, 429--436.




\bibitem{Efron} Efron, B., Hastie, T., Johnstone, I. and Tibshirani, R. (2004). Least angle regression. \textit{Annals of Statistics}, \textbf{32}, 407--499.


\bibitem{Frank} Frank, I. and Friedman, J. (1993). A statistical view of some chemometrics regression tools. \textit{Technometrics}, \textbf{35}, 109--135. 


\bibitem{Friedman1} Friedman, J., Hastie, T., H$\ddot{\textrm{o}}$fling, H. and Tibshirani, R. (2007). Pathwise coordinate optimization. \textit{Annals of Applied Statistics}, \textbf{1}, 302--332.


\bibitem{Friedman2} Friedman, J., Hastie, T. and Tibshirani, R. (2010). Regularization paths for generalized linear models via coordinate descent. \textit{Journal of Statistical Software},  \textbf{33}, 1--22.


\bibitem{Hartnett} Hartnett, M. K., Lightbody, G. and Irwin, G. W. (1998). Dynamic inferential estimation using principal components regression (PCR). \textit{Chemometrics and Intelligent Laboratory Systems},  \textbf{40}, 215--224.


\bibitem{Hastie} Hastie, T., Tibshirani, R. and Friedman, J. (2009). \textit{The Elements of Statistical Learning (2nd ed.)}. Springer, New York. 


\bibitem{hirose} Hirose, K. and Yamamoto, M. (2015). Sparse estimation via nonconcave penalized likelihood in a factor analysis model. \textit{Statistics and Computing}, \textbf{25}, 863--875.


\bibitem{kawano} Kawano, S., Fujisawa, H., Takada, T. and Shiroishi, T. (2015). Sparse principal component regression with adaptive loading. \textit{Computational Statistics \& Data Analysis}, \textbf{89}, 192--203.


\bibitem{jennrich2006rotation} Jennrich, R. I.  (2006). Rotation to simple loadings using component loss functions: The oblique case. \textit{Psychometrika}, \textbf{71}, 173--191.


\bibitem{Jolliffe3} Jolliffe, I. T. (1982). A note on the use of principal components in regression.  {\it Applied Statistics}, {\bf 31}, 300--303.


\bibitem{Jolliffe1} Jolliffe, I. T. (2002). \textit{Principal Component Analysis (2nd ed.)}.  Springer, New York. 














\bibitem{Massy} Massy, W. F. (1965). Principal components regression in explanatory statistical research. \textit{Journal of American Statistical Association},  \textbf{60}, 234--256.


\bibitem{McCullagh} McCullagh, P. and Nelder, J. A. (1989). \textit{Generalized Linear Models}. CRC press.


\bibitem{Pearson} Pearson, K. (1901). On lines and planes of closest fit to systems of points in space. \textit{Philosophical Magazine}, \textbf{2}, 559--572. 


\bibitem{RCore} R Core Team (2016). R: A language and environment for statistical computing. R Foundation for Statistical Computing, Vienna, Austria. URL https://www.R-project.org/.


\bibitem{reiss} Reiss, P. T. and Ogden, R. T. (2007). Functional principal component regression and functional partial least squares. \textit{Journal of American Statistical Association},  \textbf{102}, 984--996.


\bibitem{rosital} Rosital, R., Girolami, M., Trejo, L. J. and Cichocki, A. (2001). Kernel PCA for feature extraction  and de-noising in non-linear regression. \textit{Neural Computing \& Applications},  \textbf{10}, 231--243.




\bibitem{takada} Takada, T., Mita, A., Maeno, A., Sakai, T., Shitara, H., Kikkawa, Y., Moriwaki. K., Yonekawa, H. and Shiroishi, T. (2008). Mouse inter-subspecific consomic strains for genetic dissection of quantitative complex traits. \textit{Genome Research}, \textbf{18}, 500--508.


\bibitem{tibshirani} Tibshirani, R. (1996). Regression shrinkage and selection via the lasso. \textit{Journal of the Royal Statistical Society Series B},  \textbf{58}, 267--288.


\bibitem{wang} Wang, K. and Abbott, D. (2008). A principal components regression approach to multilocus genetic association studies. \textit{Genetic Epidemiology}, \textbf{32}, 108--118.


\bibitem{wold} Wold, H. (1975). Soft modeling by latent variables: The nonlinear iterative partial least squares approach. \textit{In Perspectives in Probability and Statistics, papers in honor of MS Bartlett}, ed. J. Gani, 520--540.


\bibitem{wu} Wu, T. T. and Lange, K. (2008). Coordinate descent algorithms for lasso penalized regression. \textit{Annals of Applied Statistics}, \textbf{2}, 224--244.




\bibitem{Zhu} Zhu, J. and Hastie, T. (2004). Classification of expression arrays by penalized logistic regression. \textit{Biostatistics}, \textbf{5}, 427--443. 


\bibitem{zou3} Zou, H. (2006). The adaptive lasso and its oracle properties. \textit{Journal of American Statistical Association},  \textbf{101}, 1418--1429.


\bibitem{zou1} Zou, H. and Hastie, T. (2005). Regularization and variable selection via the elastic net. \textit{Journal of the Royal Statistical Society Series B},  \textbf{67}, 301--320.


\bibitem{zou2} Zou, H., Hastie, T. and Tibshirani, R. (2006). Sparse principal component analysis. \textit{Journal of Computational and Graphical Statistics},  \textbf{15}, 265--286.

\end{thebibliography}
\end{document}